\documentclass[conference]{IEEEtran}
\IEEEoverridecommandlockouts
\usepackage{cite}
\usepackage{amsmath,amssymb,amsfonts}
\usepackage{algorithmic}
\usepackage{graphicx}
\usepackage{textcomp}
\usepackage{tabularx}
\usepackage{placeins}
\usepackage[table,xcdraw]{xcolor}
\usepackage{multirow}
\usepackage{amsthm}
\usepackage{url}
\usepackage{hyperref}
\usepackage{comment}
\usepackage{algorithm}
\usepackage{tikz}
\usepackage{pgfplots}
\usepackage{booktabs}
\graphicspath{ {./figure/} }
\DeclareMathSymbol{\mathbbE}{\mathord}{AMSb}{"45}
\DeclareMathOperator*{\argmin}{arg\,min}
\newcommand{\ex}{\mathbbE}
\newtheorem{proposition}{Proposition}

\def\BibTeX{{\rm B\kern-.05em{\sc i\kern-.025em b}\kern-.08em
    T\kern-.1667em\lower.7ex\hbox{E}\kern-.125emX}}
\begin{document}

\title{PureDiffusion: Using Backdoor to Counter Backdoor in Generative Diffusion Models.
	\thanks{T. V. Truong and L. B. Le are with INRS, University of Qu\'{e}bec, Montr\'{e}al, QC H5A 1K6, Canada (email: tuan.vu.truong@inrs.ca, long.le@inrs.ca).}
 \vspace{-3mm}
}

\author{\IEEEauthorblockN{Vu Tuan Truong and Long Bao Le}
\vspace{-15mm}
% \IEEEauthorblockA{
% \textit{INRS-EMT, University of Québec}\\
% Montréal, Québec, Canada \\
% tuan.vu.truong@inrs.ca, long.le@inrs.ca}
}

\maketitle

\begin{abstract}
Diffusion models (DMs) are advanced deep learning models that achieved state-of-the-art capability on a wide range of generative tasks. However, recent studies have shown their vulnerability regarding backdoor attacks, in which backdoored DMs consistently generate a designated result (e.g., a harmful image) called backdoor target when the models' input contains a backdoor trigger. Although various backdoor techniques have been investigated to attack DMs, defense methods against these threats are still limited and underexplored, especially in inverting the backdoor trigger. In this paper, we introduce PureDiffusion, a novel backdoor defense framework that can efficiently detect backdoor attacks by inverting backdoor triggers embedded in DMs. Our extensive experiments on various trigger-target pairs show that PureDiffusion outperforms existing defense methods with a large gap in terms of fidelity (i.e., how much the inverted trigger resembles the original trigger) and backdoor success rate (i.e., the rate that the inverted trigger leads to the corresponding backdoor target). Notably, in certain cases, backdoor triggers inverted by PureDiffusion even achieve higher attack success rate than the original triggers.
\end{abstract}


\section{Introduction} \label{introduction}
In recent years, diffusion models (DMs) have emerged rapidly and set the new state-of-the-art among different categories of deep generative models. By employing a multi-step denoising approach\cite{ho2020denoising}, DMs achieved superior capability in various generative tasks such as image generation\cite{rombach2022high} and 3D synthesis\cite{truong2024text}, outperforming previous methods like generative adversarial networks (GANs) and variational autoencouders (VAEs).

However, recent studies have shown that DMs are vulnerable to a wide range of attacks\cite{truong2024attacks}, among which backdoor attack is especially harmful due to its stealthiness. Specifically, to conduct a backdoor attack on DMs, attackers modify the training objective of DMs and poison the training data such that the backdoored DMs would generate a designated result (e.g., a violent image) when a trigger is stamped in the models' input. Otherwise, backdoored DMs still generate normal samples if the trigger is not activated, making it imperceptible by human observation. Moreover, the consequence might be exacerbated further if backdoored DMs are used in security-centric applications (e.g., adversarial purification) or safety-critical downstream tasks (e.g., medical imaging).

Despite these threats, defense methods for DM-targeted backdoor attacks were underexplored. Most existing backdoor detection frameworks are investigated for traditional classification models, which are inapplicable for DMs due to their differences in prediction objective and model operation. Typically, backdoor defense methods for DMs often include three main stages. The first stage is trigger inversion, in which one tries to find one or some candidate backdoor triggers from the suspicious DMs. In the second stage, backdoor detection is conducted by using the inverted triggers to figure out whether the DMs were truly backdoored or not. Once the DMs are marked as backdoor, one can choose to omit the manipulated models, or implement the third stage called backdoor removal, which eliminates the backdoor effect embedded in the target DMs. Among these stages, trigger inversion is often the most challenging task as the knowledge about the suspicious models is still limited at this stage. It is also the most impactful stage since the inverted trigger's quality greatly decides backdoor detection performance. In this paper, we introduce PureDiffusion, a backdoor defense framework that focuses on high-quality trigger inversion.

Prior to PureDiffusion, existing backdoor defense methods for DMs either ignore the trigger inversion stage or can invert only low-quality trigger. For instance, both DisDet\cite{sui2024disdet} and UFID\cite{guan2024ufid} assume that they already have a set of candidate triggers, and the two methods are designed to verify which of the candidates is the true trigger. DisDet\cite{sui2024disdet} relies on the following observation: If the model's input contains the trigger, it will consistently result in a particular image (i.e., the backdoor target), otherwise the generated results should be very diverse. Thus, the authors compute the cosine similarity between every pairs of generated image, constructing a similarity graph to assess each candidate trigger. If the similarity of a candidate exceeds a predefined threshold, it is considered a backdoor trigger. On the other hand, the authors of UFID\cite{guan2024ufid} argue that the trigger's distribution should be significantly different from a Gaussian noise's distribution. Thus, they computes the Kullback–Leibler (KL) divergence between candidate triggers and a standard Gaussian noise, then choose a statistical threshold to determine which one is the true trigger. However, it is impractical to assume that the set of candidate triggers is available. To the best of our knowledge, only the work Elijah\cite{an2024elijah} offers a practical method for trigger inversion. Elijah finds the trigger based on the distribution shift caused by the trigger in each diffusion step. However, it only computes the shift in the last timestep and heuristically chooses a scale of 0.5 for the shift without any insighful justification. Consequently, Elijah can invert only simple triggers, while the inverted triggers are often of low quality. When we modify the trigger by increasing its size or altering its shape, the performance of Elijah's trigger inversion decreases rapidly.

In our work PureDiffusion, we present a practical method to compute the scale of trigger-related distribution shift for every denoising step, proved by both empirical and theoretical analyses. Based on the computed shift scales, we can use gradient descent to learn the trigger over multiple timesteps, improving the capability of trigger inversion. The contribution of our paper can be summarized as follows:
\begin{itemize}
    \item We introduce PureDiffusion, a backdoor defense framework for DMs based on a multi-timestep trigger inversion method that outperforms existing work by a large gap.
    \item We propose the first solution to estimate the scale of trigger-related distribution shift for all denoising steps, making it feasible to invert the trigger based on multiple diffusion timesteps.
    \item We conduct extensive experiments on various trigger-target pairs to showcase the efficiency of PureDiffusion. Its performance is evaluated based on different metrics such as uniform score, trigger fidelity, and sampling quality, showing the outstanding capability of PureDiffusion compared to existing work.
\end{itemize}

The rest of the paper is presented as follows. Section \ref{section:background} provides necessary background knowledge in terms of DMs and DM-targeted backdoor attacks. Section \ref{section:method} describes our method PureDiffusion. Section \ref{section:experiment} presents a wide range of experiments to illustrate the efficiency of PureDiffusion, while section \ref{section:conclusion} concludes the paper.

\section{Background Knowledge}\label{section:background}

\subsection{Diffusion Models}
DMs are trained to generate images from Gaussian noise via two main processes, a forward (diffusion) process and a reverse (denoising) process. In the forward process, a small amount of Gaussian noise is iteratively added to the training images in $T$ steps until they are totally destroyed into a standard Gaussian distribution at the final step. In the reverse process, a deep neural network model $\theta$ is trained to reconstruct these images by removing the added noise in $T$ denoising steps. While there are different categories of DMs, they all follow the above operation. In this paper, we primarily focus on denoising diffusion probabilistic model (DDPM), in which the diffusion process is modelled as a Markovian chain with $T$ states corresponding to $T$ diffusion steps, noising a clean image $x_0$ into a Gaussian noise $x_T$. Formally, the transition between two consecutive forward steps is:
\begin{equation}
\label{equation:forward-transition}
    q(x_t|x_{t-1}) = \mathcal{N}(x_t;\sqrt{1-\beta_t}x_{t-1}, \beta_t\mathbf{I}),
\end{equation}
where $\beta_t \in (0,1)$ is the scale of noise at step $t$. By applying the reparameterization trick $T$ times on equation~\ref{equation:forward-transition}, we obtain the following property that enables direct sampling of $x_t$ from the clear image $x_0$:
\begin{equation}
\label{equation:direct-sampling}
    x_t = \sqrt{\Bar{\alpha}_t}x_0 + \sqrt{1-\Bar{\alpha}_t}\epsilon_0,
\end{equation}
where $\Bar{\alpha}_t = \prod_{i=1}^{t} \alpha_{i}$ and $\alpha_t = 1-\beta_t$.

The reverse process employs a deep neural network parameterized by $\theta$ to predict the noise in each step, based on the following denoising transition:
\begin{equation}
\label{equation:reverse-transition}
    p_\theta(x_{t-1}|x_t) = \mathcal{N}(x_{t-1}; \mu_\theta(x_t,t), \Sigma_\theta(x_t,t)),
\end{equation}
where  $\mu_\theta(x_t,t)$ and $\Sigma_\theta(x_t,t)$ are the mean and variance of image distribution at step $t$, predicted by using the neural network $\theta$. The loss function to train this neural network is the KL divergence between the predicted $p_\theta(x_{t-1}|x_t)$ and the ground-truth posterior $ q(x_{t-1}|x_t,x_0)$ derived based on (\ref{equation:direct-sampling}). As a result, the final loss function can be simplified into the following form:
\begin{equation}
\label{equation:loss-ddpm}
    \mathcal{L} = \ex_{x_0,\epsilon} \left[ \left\| \epsilon - \epsilon_\theta(x_t,t) \right\|^2_2 \right],
\end{equation}
where $\epsilon \sim \mathcal{N}(0,\mathbf{I})$ is a sampled noise, and $\epsilon_\theta(x_t,t)$ is the noise predicted by the neural network. 
After training on a specific dataset, the neural network $\theta$ can be used to generate image samples based on the following reserve process:
\begin{equation}
\label{equation:ddpm-sampling}
    x_{t-1} = \frac{1}{\sqrt{\alpha_t}} \left( x_t - \frac{1-\alpha_t}{\sqrt{1-\Bar{\alpha}_t}}\epsilon_\theta(x_t,t) \right) + \sigma_t\epsilon,
\end{equation}
where $\sigma_t = \frac{(1-\alpha_t)(1-\Bar{\alpha}_{t-1})}{1-\Bar{\alpha}_{t}}$. 
% To obtain a new generated image $x_0$, we sample $T$ times from a random Gaussian distribution $x_T \sim \mathcal{N}(0,\mathbf{I})$ based on the above equation (\ref{equation:ddpm-sampling}).

\subsection{Backdoor Attacks on Diffusion Models}
The goal of backdoor attacks on a DM is to make it generates a designated backdoor target $x^*_0$ when a backdoor trigger $\delta$ is stamped into the DM's input noise. In this case, the DM's input is $x^*_T \sim \mathcal{N}(\delta,\mathbf{I})$. Note that we use ``$*$" to denote the backdoor case. To do so, the forward transition is modified to gradually add a small amount of the backdoor trigger $\delta$ into the image distribution, resulting in the following backdoor forward transition:
\begin{equation}
\label{equation:backdoor-forward}
    q(x^*_t|x^*_{t-1}) = \mathcal{N}(x^*_t;\sqrt{1-\beta_t}x^*_{t-1} + m(t)\delta, n(t)\mathbf{I}),
\end{equation}
where $m(t)$ and $n(t)$ are functions that determine the amount of the trigger and noise added at each step, respectively. It can be seen that if $m(t)=0$ and $n(t)=\beta_t$, the equation (\ref{equation:backdoor-forward}) will have the same form with the equation (\ref{equation:forward-transition}), in which there is no trigger added to the image distribution. 

Different backdoor methods use different $m(t)$ and $n(t)$, but these functions must be chosen carefully to make the DMs trainable. For example, BadDiffusion\cite{Chou2023CVPR} chooses $m(t)=1-\sqrt{\Bar{\alpha}_t}$ and $n(t)=\beta_t$, which results in the following property:
\begin{equation}
\label{equation:baddiffusion-direct-sampling}
    x^*_t = \sqrt{\Bar{\alpha}_t}x^*_0 + (1-\sqrt{\Bar{\alpha}_t})\delta + \sqrt{1-\Bar{\alpha}_t}\epsilon,
\end{equation}

On the other hand, TrojDiff\cite{chen2023trojdiff} chooses $m(t)=k_t(1-\gamma)$ and $n(t)=\gamma\beta_t$, where $\gamma \in [0,1]$ is a predefined blending coefficient, and $k = \{k_0, k_1,...,k_t,...,k_T\}$ is intentionally selected to offer the direct sampling from $x^*_0$ to $x^*_t$:
\begin{equation}
\label{equation:trojdiff-direct-sampling}
    x^*_t = \sqrt{\Bar{\alpha}_t}x^*_0 + \sqrt{1-\Bar{\alpha}_t}(1-\gamma)\delta + \sqrt{1-\Bar{\alpha}_t}\gamma\epsilon.
\end{equation}

Based on the above backdoored forward process, the reverse process and training objective can be derived accordingly. During training, a certain proportion of the training data (5-20\%) is poisoned by the trigger and target images, combined with the backdoor loss function to attack the targeted DMs.

\section{PureDiffusion: Methodology}\label{section:method}
PureDiffusion is designed to invert backdoor triggers from manipulated DMs. To do so, we first show that a backdoored forward process will consistently add a trigger-related distribution shift into the image distribution. Based on this property, we propose a practical method to figure out the scale of distribution shift in every timestep of the diffusion processes. Utilizing the computed scales, we use gradient descent to learn the triggers from multiple timesteps, resulting in high-quality inverted triggers.

\subsection{Trigger-Related Distribution Shift}
Applying reparameterization on the backdoored forward transition in equation (\ref{equation:backdoor-forward}), the transition can be viewed as:
\begin{equation}
\label{equation:backdoor-reparameter}
    x^*_t = \sqrt{1-\beta_t}x^*_{t-1} + \underbrace{m(t)\delta}_\textrm{trigger shift} + \underbrace{n(t)\epsilon}_\textrm{noise shift}.
\end{equation}

\begin{figure}[h!]
	\centering
	\includegraphics[scale=0.145]{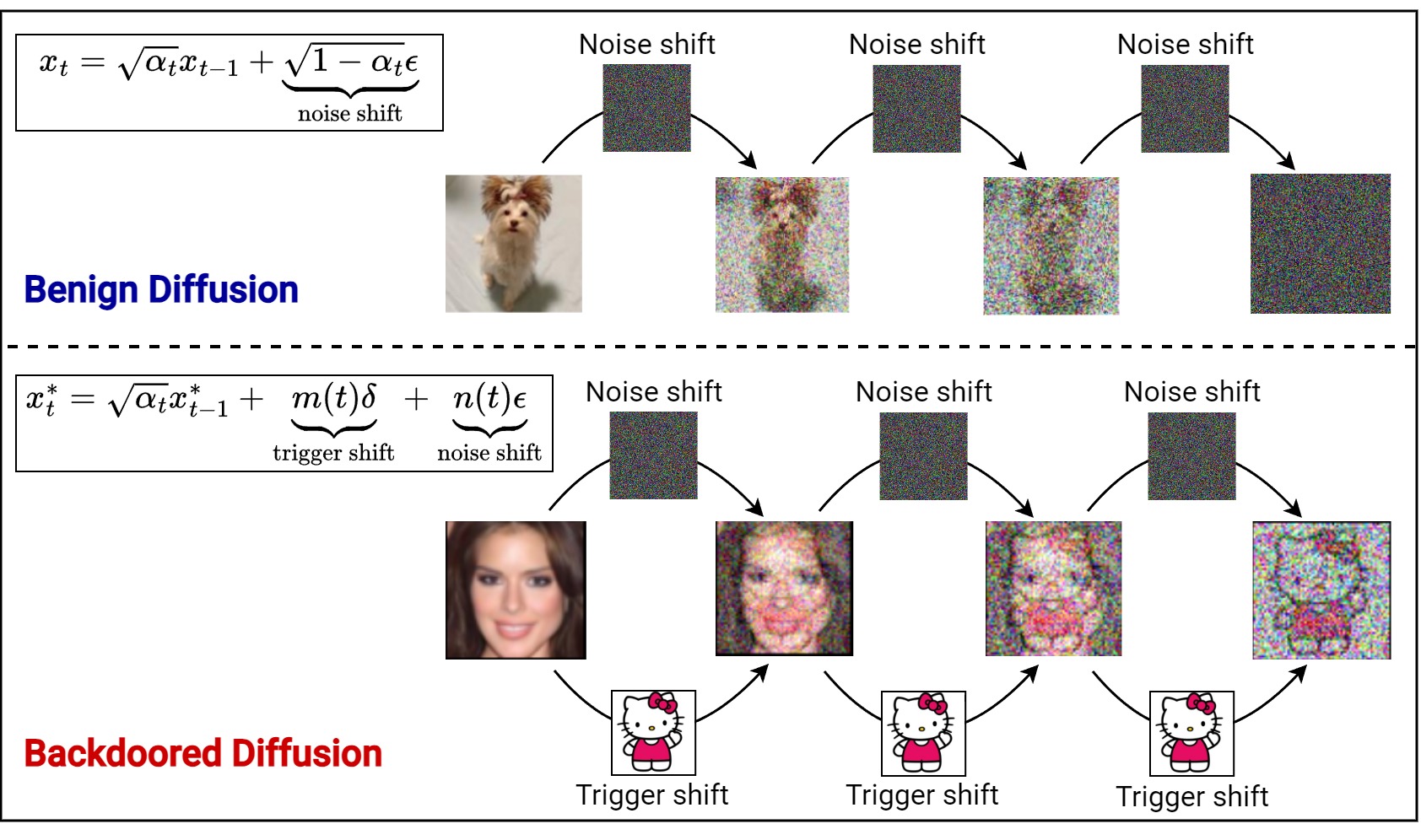}
	\caption{A visualization of benign and backdoored diffusion processes from the view of distribution shift.}
	\label{figure:backdoor-vs-normal}
\end{figure}

Intuitively, the ``noise shift" term adds a small amount of Gaussian noise into the image distribution in each timestep, destroying the image gradually over time. This term is included in both benign and backdoored DMs. On the other hand, the ``trigger shift" term will iteratively shift the image distribution towards the trigger distribution. Consequently, the final result at step $T$ is a noisy trigger image $x^*_T \sim \mathcal{N}(\delta,\mathbf{I})$. Fig.~\ref{figure:backdoor-vs-normal} visualizes the distribution shifts of benign and backdoored DMs over timesteps.

Since the neural network $\theta$ is trained to reverse the forward process, the backdoored denoising process must preserve the trigger shift in every timestep, but in a reverse direction, to reconstruct the backdoor target from the noisy trigger. If the scale of the trigger shift in the reverse process is known for every timestep, we can learn the backdoor trigger via gradient descent\cite{an2024elijah}. However, the scale of the trigger shift predicted by the neural network is different than those in the forward process due to the difference in coefficients, making trigger inversion challenging. An existing work named Elijah\cite{an2024elijah} heuristically chooses a scale of $0.5$ for the first denoising step $T$, and learn the trigger on only that timestep. As a result, although Elijah can invert some simple triggers, the inverted triggers are often of low quality, and it failed in inverting more sophisticated trigger shapes.

\subsection{Analysis of Trigger Shift in Reverse Process}
In this section, we introduce a feasible solution to compute the scale of trigger shift predicted by the neural network $\theta$ in every timestep, which is unsolved in existing work. Due to the limited scope of the paper, we opt to analyze BadDiffusion\cite{Chou2023CVPR}, while similar analyses can be used for other backdoor methods like TrojDiff\cite{chen2023trojdiff} and VillanDiffusion\cite{chou2024villandiffusion}, which are similar but come with different $m(t)$ and $n(t)$.

\subsubsection{Theoretical Analysis} Let $\lambda_t$ denotes the scale of trigger shift in timestep $t$, and all trigger shift scales are in $\lambda = \{ \lambda_0, \lambda_1,...,\lambda_t,...,\lambda_T \}$, we propose the following proposition.

\begin{proposition} The trigger shift scales $\lambda$ is the same between different backdoor triggers, regardless of their shape and size.
\end{proposition} 
\label{proposition:trigger-scale}

\begin{proof}
In the backdoored reverse process, the equation (\ref{equation:ddpm-sampling}) can be expressed as:
\begin{equation}
\label{equation:ddpm-sampling-proportional}
    x^*_{t-1} = \frac{x^*_t}{\sqrt{\alpha_t}} - \frac{1-\alpha_t}{\sqrt{\alpha_t}\sqrt{1-\Bar{\alpha}_t}}\epsilon_\theta(x^*_t,t) + \sigma_t\epsilon.
\end{equation}

Here, the predicted noise $\epsilon_\theta(x^*_t,t)$ contains both the reverse trigger shift and the noise shift. However, the scale of trigger shift is intractable as it is from the neural network's output. 

By applying the Bayes' rule on the equation (\ref{equation:backdoor-forward}) with $m(t)=1-\sqrt{\Bar{\alpha}_t}$ and $n(t)=\beta_t$, we obtain the posterior of BadDiffusion's forward process, which has been shown in \cite{Chou2023CVPR}:
\begin{equation}
\label{equation:baddiffusion-posterior}
    x^*_{t-1} = \frac{x^*_t}{\sqrt{\alpha_t}} - \frac{1-\sqrt{\alpha_t}}{\sqrt{\alpha_t}}\delta + \sigma_t\epsilon
\end{equation}

Subtracting (\ref{equation:ddpm-sampling-proportional}) from (\ref{equation:baddiffusion-posterior}), we obtain:
\begin{align}
    \epsilon_\theta(x^*_t,t) &= \frac{(1-\sqrt{\alpha_t})\sqrt{1-\Bar{\alpha}_t}}{1-\alpha_t}\delta \label{equation:noise-scale} \\
    &= \lambda_t\delta \label{equation:noise-scale-lambda}.
\end{align}
% \begin{equation}
% \label{equation:noise-scale}
%     \epsilon_\theta(x^*_t,t) = \frac{(1-\sqrt{\alpha_t})\sqrt{1-\Bar{\alpha}_t}}{1-\alpha_t}\delta = \lambda_t\delta.
% \end{equation}

As $\lambda_t=\frac{(1-\sqrt{\alpha_t})\sqrt{1-\Bar{\alpha}_t}}{1-\alpha_t}$ only depends on the fixed noise schedule $\alpha_t$ without relying on the trigger $\delta$, the Proposition is proved.
\end{proof}

\subsubsection{Empirical Analysis}
The Proposition 1 indicates that if we can find the trigger-shift scales of an arbitrary trigger (which is not the true trigger), we can treat it as the trigger-shift scales of such the true trigger.
Thus, to extract these scales, we first choose an surrogate trigger $\hat{\delta}$, then simulate a backdoor attack targeting on a copy version the suspicious DM. Using this double-backdoored DM, we can compute the predicted noise $\epsilon_\theta(x^*_t,t)$ in every timestep by running the backdoor reverse process using the trigger $\hat{\delta}$. Since both $\epsilon_\theta(x^*_t,t)$ and $\hat{\delta}$ are known, from (\ref{equation:noise-scale-lambda}), we can find every $\lambda_t$ via the following minimization problem:
\begin{equation}
\label{equation:noise-scale-minimize}
    \lambda_t = \argmin_{\lambda_t} \left\| \epsilon_\theta(x^*_t,t) - \lambda_t\hat{\delta}  \right\|^2_2.
\end{equation}

This minimization problem can be solved by setting the derivative to zero to find the optimal $\lambda_t$, resulting in:
\begin{equation}
\label{equation:noise-scale-replicate}
    \lambda_t = \frac{\epsilon_\theta(x^*_t,t)\cdot\hat{\delta}}{|| \hat{\delta} ||^2_2}.
\end{equation}

Because $\lambda = \{\lambda_0, \lambda_1,...\lambda_T\}$ are the trigger-shift scales of the surrogate trigger $\hat{\delta}$, it is also valid for the true trigger $\delta$ although the two triggers are different from each other (according to the Proposition 1).

\begin{figure}[h!]
	\centering
	\includegraphics[scale=0.45]{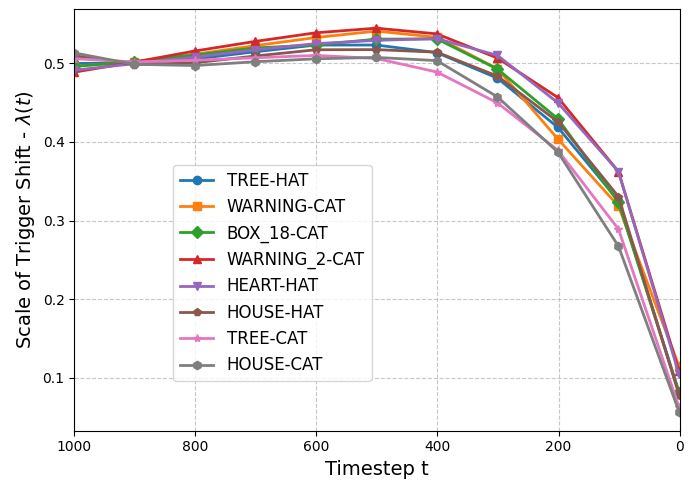}
	\caption{Visualization of the trigger shift's scale $\lambda_t$ with different trigger-image pairs and $T=1000$.}
	\label{figure:distribution_shift}
\end{figure}

Fig.~\ref{figure:distribution_shift} shows the value of $\lambda$ computed by the proposed method, using different trigger-target pairs. It can be seen that the trigger-shift scales $\lambda$ only vary slightly between the experimented pairs due to instinctive stochastic factors of DDPMs; but in overall they are still in the same form, which empirically proves the Proposition 1. 
% The average scale at the timestep $T$ is around $0.5$, which shows why Elijah\cite{an2024elijah} is efficient with this coefficient.

\subsection{Multi-Timestep Trigger Inversion for Backdoor Detection}
Once successfully extracting the trigger-shift scales $\lambda$ via the proposed method, from the equation (\ref{equation:noise-scale-lambda}), we can find the true trigger $\delta$ based on the following optimization problem: 
\begin{equation}
\label{equation:trigger-minimize}
    \delta = \argmin_{\delta} \ex_{\epsilon} \left\| \epsilon_\theta(x^*_t(\delta,\epsilon),t) - \lambda_t\delta  \right\|^2_2
\end{equation}
\begin{equation}
\label{equation:trigger-loss}
    Loss = \ex_{\epsilon} \left\| \epsilon_\theta(x^*_t(\delta,\epsilon),t) - \lambda_t\delta  \right\|^2_2.
\end{equation}

Here,  $\epsilon_\theta(x^*_t(\delta,\epsilon),t)$ is the neural network's output (i.e., predicted noise) at timestep $t$, given that its input at the first denoising step contains the trigger $\delta$. As a result, we can use gradient descent to learn $\delta$, with $t$ is sampled randomly from 0 to $T$. For each sampled timestep $t$, the backpropagation flow passes through the neural network $T-t$ times, from the first denoising step $T$ to the timestep $t$. This ensures that the denoising model can consistently keep the learned trigger shift over different timesteps. Algorithm~\ref{algorithm:trigger-inversion} summarizes our trigger inversion method, with $\epsilon_\theta(x^*_t(\delta,\epsilon),t)$ is simplified to $\epsilon_\theta(t)$ for the ease of reading. Thus, please note that $\epsilon_\theta(t)$, $\epsilon_\theta(x^*_t,t)$, and $\epsilon_\theta(x^*_t(\delta,\epsilon),t)$ are the same in this paper.

\begin{algorithm}
\small
\caption{PureDiffusion: Backdoor Trigger Inversion}\label{algorithm:trigger-inversion}
\textbf{Input:} Suspicious model $\theta$, number of training epochs $N$ \\
\textbf{Output:} Predicted backdoor trigger $\delta$

\begin{algorithmic}[1]
\STATE Randomize an arbitrary trigger $\hat{\delta}$.
\STATE Use $\hat{\delta}$ to backdoor the suspicious model $\theta$.
\STATE // \textit{Estimating the trigger shift scales}
\STATE Initialize a list of trigger shift scales $\lambda = \{\lambda_0, \lambda_1,...,\lambda_T \}$.
\FOR{$t = T, T-1,...,1$} 
    \STATE Predict the noise $\epsilon_\theta(t)$ using $\hat{\delta}$.
    \STATE $\lambda_t = (\epsilon_\theta(t)\cdot\hat{\delta}) / || \hat{\delta} ||^2_2$ \space (\ref{equation:noise-scale-replicate}).
\ENDFOR
\STATE // \textit{Learn the true trigger}
\STATE Initialize a learnable trigger $\delta$ which requires gradient.
\FOR{$epoch = 0, 1,...,N$} 
    \STATE Sample a random timestep $t \in [0, T]$.
    \STATE Predict the noise $\epsilon_\theta(t)$ using $\delta$.
    \STATE $Loss = \ex_{\epsilon} \left\| \epsilon_\theta(t) - \lambda_t\delta  \right\|^2_2$ \space (\ref{equation:trigger-loss}).
    \STATE Backprop the computed loss to optimize $\delta$.
\ENDFOR
\RETURN $\delta$
\end{algorithmic}
\end{algorithm}

Once successfully revert the trigger, it is easy to verify whether the DM was backdoored or not. For instance, we can use the uniform score presented in Section~\ref{section:experiment} as a mean of backdoor detection. If the suspicious DM was backdoored, the inverted trigger should make the model consistently produces the backdoor target in different trials, resulting in a low uniform score. Otherwise, if the model is not backdoored, the uniform score must be high as the DM generates diverse images among different trials due to its stochastic factors. By choosing a predefined threshold, we can easily conduct backdoor detection. Obviously, the performance of backdoor detection mainly depends on the quality of the inverted trigger. Thus, we primarily focus on trigger inversion, while backdoor detection is not presented in details in this paper due to the limited scope.

\begin{table*}[]
\centering
\caption{Performance of triggers inverted by PureDiffusion compared to Elijah's triggers and ground-truth triggers.}
\label{table:performance}
\begin{tabular}{@{}|lclccc|lclccc|@{}}
\toprule
\multicolumn{1}{|c}{\textbf{Trigger}} & \textbf{Target} & \multicolumn{1}{c}{\textbf{Method}} & \textbf{Uniform $\downarrow$} & \textbf{ASR $\uparrow$} & \textbf{L2 dist $\downarrow$} & \multicolumn{1}{c}{\textbf{Trigger}} & \textbf{Target} & \multicolumn{1}{c}{\textbf{Method}} & \textbf{Uniform $\downarrow$} & \textbf{ASR $\uparrow$} & \textbf{L2 dist $\downarrow$} \\ \midrule
\multirow{3}{*}{Stop 14} & \multirow{3}{*}{Hat} & GT & 0.02487 & 87.5\% & 0 & \multirow{3}{*}{Tree} & \multirow{3}{*}{Cat} & GT & 0.00123 & 100\% & 0 \\
 &  & Elijah & 0.41517 & 50\% & 39.6534 &  &  & Elijah & 0.40011 & 25\% & 41.9048 \\
 &  & PureDiff & 0.00061 & 87.5\% & 38.7664 &  &  & PureDiff & 0.00075 & 100\% & 34.0282 \\ \midrule
\multirow{3}{*}{Stop 18} & \multirow{3}{*}{Hat} & GT & 0.00135 & 75\% & 0 & \multirow{3}{*}{House} & \multirow{3}{*}{Cat} & GT & 0.00114 & 100\% & 0 \\
 &  & Elijah & X & 0\% & 39.7290 &  &  & Elijah & 0.00078 & 100\% & 28.8905 \\
 &  & PureDiff & 0.42035 & 68.75\% & 33.5083 &  &  & PureDiff & 0.00065 & 100\% & 26.8641 \\ \midrule
\multirow{3}{*}{Heart} & \multirow{3}{*}{Hat} & GT & 0.00184 & 87.5\% & 0 & \multirow{3}{*}{Warning1} & \multirow{3}{*}{Cat} & GT & 0.28175 & 62.5\% & 0 \\
 &  & Elijah & 0.33882 & 56.25\% & 33.6720 &  &  & Elijah & 0.00071 & 93.75\% & 29.8783 \\
 &  & PureDiff & 0.13693 & 68.75\% & 31.1835 &  &  & PureDiff & 0.00068 & 100\% & 31.7135 \\ \midrule
\multirow{3}{*}{Tree} & \multirow{3}{*}{Hat} & GT & 0.00177 & 100\% & 0 & \multirow{3}{*}{Warning2} & \multirow{3}{*}{Cat} & GT & 0.03914 & 87.5\% & 0 \\
 &  & Elijah & 0.34878 & 56.25\% & 35.7612 &  &  & Elijah & 0.31639 & 12.5\% & 34.4414 \\
 &  & PureDiff & 0.00075 & 87.5\% & 30.5990 &  &  & PureDiff & 0.00108 & 87.5\% & 31.5722 \\ \midrule
\multirow{3}{*}{Warning1} & \multirow{3}{*}{Hat} & GT & 0.00525 & 87.5\% & 0 & \multirow{3}{*}{House} & \multirow{3}{*}{Hat} & GT & 0.00082 & 100\% & 0 \\
 &  & Elijah & 0.30535 & 18.75\% & 35.6798 &  &  & Elijah & 0.00107 & 87.5\% & 32.1612 \\
 &  & PureDiff & 0.00085 & 87.5\% & 34.1811 &  &  & PureDiff & 0.00074 & 100\% & 27.5943 \\ \bottomrule
\end{tabular}
\end{table*}

\begin{figure}[h!]
	\centering
	\includegraphics[scale=0.16]{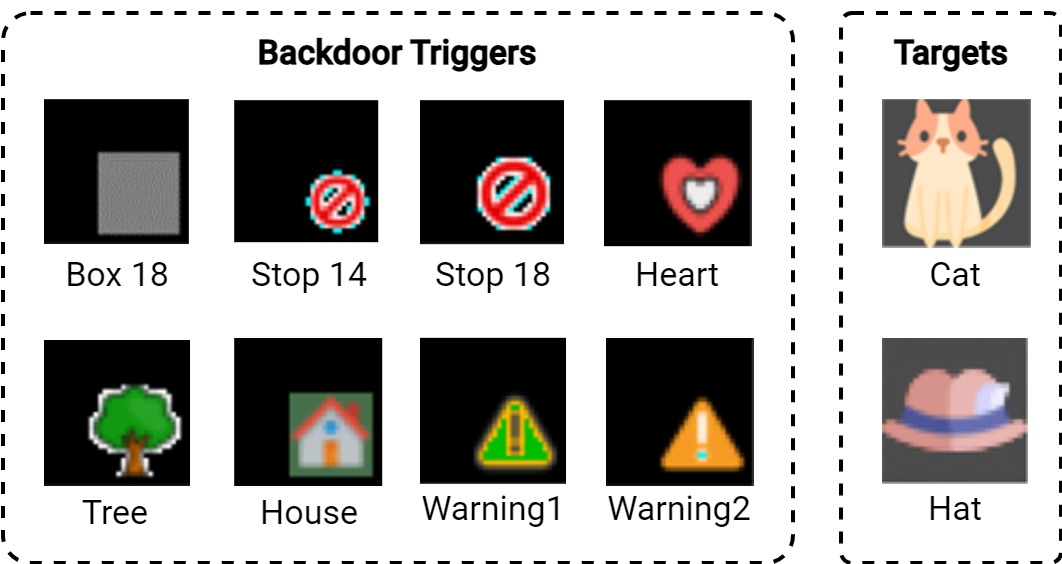}
	\caption{Visualization of different backdoor triggers and targets used in our experiments. The number 14 or 18 indicates the trigger size in pixels.}
	\label{figure:trigger-target}
\end{figure}

\section{Experimental Results}\label{section:experiment}

\subsection{Experimental Setup}

\subsubsection{Dataset and Attack Settings}
In our experiments, we use DDPM with $T=1000$, while the noise schedule starts from $\beta_0=0.0001$ to $\beta_T=0.02$. Due to the limited scope of the paper, we mainly evaluate the performance of our defense framework against BadDiffusion\cite{Chou2023CVPR}, while it can be adopted easily for other backdoor methods such as TrojDiff\cite{chen2023trojdiff} and VillanDiffusion\cite{chou2024villandiffusion}. The experiments are performed on CIFAR-10 dataset\cite{krizhevsky2009learning}

\subsubsection{Evaluation Metrics}
To assess the performance of PureDiffusion, we use the following three metrics:
\begin{itemize}
    \item \textbf{Distance from the ground-truth trigger:} This is the L2 distance between the inverted trigger and the ground-truth trigger. The lower distance, the higher trigger quality (as the inverted trigger looks more like the ground-truth one).
    \item \textbf{Attack success rate of inverted trigger:} We feed the inverted trigger as input of the suspicious DM and observe if its output is the backdoor target. If the distance between the output image and the backdoor target is smaller than a predefined threshold, it is considered a successful attack. We repeat this $N$ times (e.g., generating 100 samples) to compute the attack success rate (ASR). The higher ASR, the higher trigger quality.
    \item \textbf{Uniform score:} To compute this score, we generate $M$ different image samples using the suspicious DM with its input contains the inverted trigger. If the inverted trigger is of high quality, these $M$ samples should resemble each other as they resemble the backdoor target. As a result, the uniform score is such the total distance of every pair of generated samples. Thus, the lower uniform score, the higher trigger quality.
    
\end{itemize}

\subsubsection{Baseline Methods}
Prior to our study, only Elijah\cite{an2024elijah} investigates trigger inversion for DMs. While our method successfully computes the trigger shift scales for all timesteps, proved by both empirical and theoretical analyses, Elijah only uses the timestep $T$ with a shift scale of 0.5 without any justification. To showcase the capability of our framework, we compare its performance with both Elijah triggers and the ground-truth triggers. Regarding Elijah, we run the experiments based on their provided source code\cite{an2024elijah}.

\subsubsection{Trigger Learning Setup}
For trigger inversion with multiple timesteps, we need to iteratively run the neural network multiple times for denoising, in which the output of the previous timestep is the input of the current timestep. We backprop the gradients through all of these timesteps to learn the trigger. Due to our limitation in computational resources, we only learn the trigger based on the first 10 steps of the denoising process (i.e., $t$ is randomly sampled from timestep $T$ to timestep $T-10$), using a batch size of 40.
In the learning settings, we use Adam optimizer with a learning rate of 0.1. The number of epochs varies depending on the difficulty of the trigger. While we use 30 epochs in most cases, some triggers may require more epochs to learn it efficiently.

\begin{figure*}[h!]
	\centering
	\includegraphics[scale=0.14]{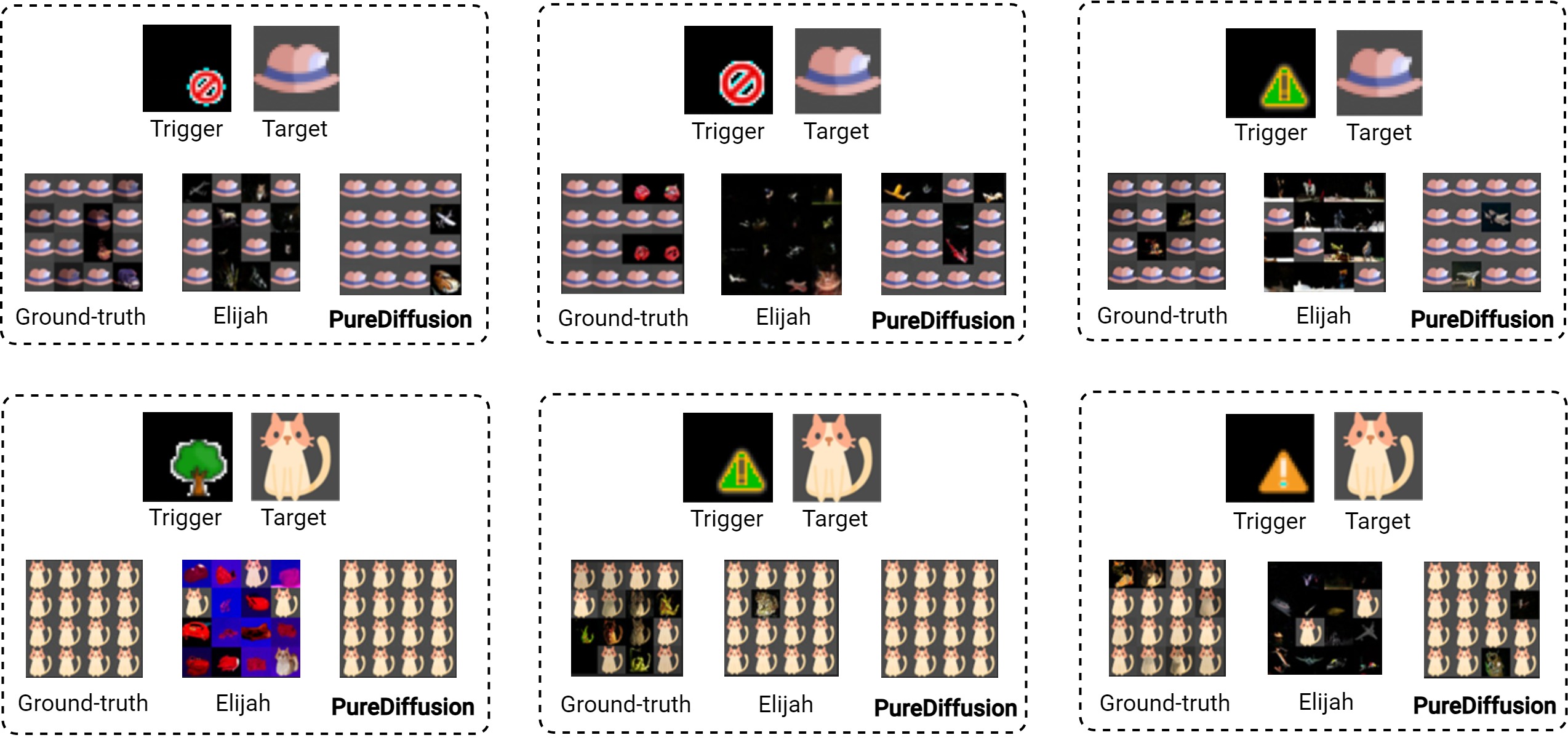}
	\caption{Sampling results of three trigger types, where triggers inverted by PureDiffusion can sample backdoor targets better than the ground-truth triggers.}
	\label{figure:perfomance-sampling}
\end{figure*}

\subsection{Performance Analysis}
Table~\ref{table:performance} presents the performance of PureDiffusion in comparison to Elijah, experimented on various trigger-target pairs. Experimented triggers and targets are visualized in Fig.~\ref{figure:trigger-target}. Through some experiments, we discovered that different trigger shapes and sizes impact greatly on the performance of trigger inverson and backdoor detection. From experimental observation, some triggers (e.g., stop sign), for some reasons, are more ``difficult" to be inverted than some others.

\subsubsection{Sampling Quality}
To evaluate the quality of image samples generated from the inverted triggers, we use both uniform score and ASR.
In terms of uniform score, PureDiffusion shows it superior performance compared to Elijah. For example, with ``Tree" as trigger and ``Cat" as target, our method achieves a uniform score of 0.00075, which is 500 times lower than which predicted by Elijah. For many other trigger-target pairs such as Warning1-Hat, Tree-Hat, and Stop14-Hat, the performance gap is also very large between the two methods. In general, PureDiffusion achieves lower uniform score in all experimented cases.

Regarding ASR, our method also shows its superior performance compared to Elijah in all experiments. For instance, in case of Warning1-Hat pair, PureDiffusion achieves 87.5\% ASR, while Elijah's inverted trigger can generate the backdoor target in only 18.75\% trials. In another difficult case with Stop18-Hat pair, Elijah's trigger cannot even generate the backdoor target in any trial, leading to an ASR of 0\%. On the other hand, PureDiffusion still achieves 68.75\% ASR in that case. Notably, the ASR of PureDiffusion's triggers is comparable with the ground-truth triggers in most cases.

\subsubsection{Trigger Fidelity} 
To asses trigger fidelity, we mainly compute l2 distance between the inverted trigger and the ground-truth one. As shown in Table~\ref{table:performance}, in all test cases, triggers inverted by PureDiffusion achieved lower l2 distance than those predicted by Elijah. For certain triggers such as ``Tree" and ``Stop18", the difference is even more significant. This proves that our method can ensure a higher level of fidelity than Elijah. Using more than 10 timesteps for backpropagation can potentially increase trigger fidelity, but it requires more computational resource. Thus, we leave it for future work.

\subsection{PureDiffusion for Reinforced Backdoor Attack}
As shown in Table~\ref{table:performance}, the uniform scores of PureDiffusion's triggers are even lower than the ground-truth triggers in more than $50\%$ of experimented cases, including Stop14-Hat, Tree-Hat, Warning1-Hat, Tree-Cat, House-Cat, Warning1-Cat, Warning2-Cat, and House-Hat. In these cases, the ASR of triggers inverted by our method is comparable or even higher than the ground-truth triggers. This implies the following insight: While PureDiffusion is initially designed as a defense method for backdoor attacks, it can be used to reinforce such backdoor attacks. By applying PureDiffusion to learn a new trigger, the learned trigger potentially offers higher ASR, while it is harder to be detected by human observers as it contains a certain amount of random noise during learning, thus improving stealthiness.

Fig.~\ref{figure:perfomance-sampling} illustrates the above observation, where PureDiffusion triggers can generate the backdoor targets better than not only Elijah, but also the ground-truth triggers. A potential reason for this interesting result is that PureDiffusion's triggers are less visually detailed than the ground-truth ones, making its backdoor patterns easier to be recognized by backdoored DMs.

\section{Conclusion} \label{section:conclusion}
In this paper, we propose PureDiffusion, a backdoor defense framework for DMs, focusing on trigger inversion. First, we propose a novel reverse-engineering method to estimate the trigger shift scales caused by the backdoor diffusion processes. This method is proved by both empirical and theoretical analyses. Second, by using the estimated trigger shift scales, we learn the true trigger over multiple denoising steps, offering high-quality trigger inversion. Experimental results have showcased the efficiency of PureDiffusion, as it outperforms existing work with a large performance gap. Furthermore, PureDiffusion's inverted triggers even outperform the ground-truth triggers, making it a potential reinforcement method for backdoor attacks.

\footnotesize{
\bibliographystyle{IEEEtran}
\bibliography{reference.bib}
}

\end{document}